\documentclass{article}
\usepackage[utf8]{inputenc}
\usepackage{amsfonts,amsmath}
\usepackage{mdframed}
\usepackage{graphicx}
\usepackage{hyperref}

\usepackage{multirow}
\usepackage{array}
\usepackage{subcaption}

\usepackage{tikz}
\usetikzlibrary{arrows.meta,,positioning}

\usepackage[labelfont=bf]{caption}
\usepackage[ruled,boxed,linesnumbered,commentsnumbered,longend]{algorithm2e}

\usepackage{amsthm}
\theoremstyle{definition}
\newtheorem{prop}{Proposition}

\usepackage{multirow}
\usepackage{breqn}

\def\old#1{}

\def\tl{\tilde}

\def\m{\mu}

\def\frac#1#2{{#1\over #2}}

\old{

}

\title{Superior Computer Chess with Model Predictive\\
 Control, Reinforcement Learning, and Rollout\footnote{
This work was carried out at the Fulton School of Computing, and Augmented Intelligence, Arizona State University, Tempe, AZ.}}

\author{Atharva Gundawar, Yuchao Li, and Dimitri Bertsekas\\
School of Computing, and Augmented Intelligence,\\ Arizona State University, Tempe, AZ.}

\usepackage{setspace}
\begin{document}
\maketitle

\begin{abstract}
In this paper we apply model predictive control (MPC), rollout, and reinforcement learning (RL) methodologies to computer chess. We  introduce a new architecture for move selection, within which available chess engines are used as components. One engine is used to provide position evaluations in an approximation in value space MPC/RL scheme, while a second  engine is used as nominal opponent, to emulate or approximate the moves of the true opponent player.

We show that our architecture improves substantially the performance of the position evaluation engine. In other words {\it our architecture provides an additional layer of intelligence, on top of the intelligence of the engines on which it is based\/}. This is true for any engine, regardless of its strength: top engines such as Stockfish and Komodo Dragon (of varying strengths), as well as weaker engines.

Structurally, our basic architecture selects moves by a one-move lookahead search, with an intermediate move generated by a nominal opponent engine, and followed by a position evaluation by another chess engine. Simpler schemes that forego the use of the nominal opponent, also perform better than the position evaluator, but not quite by as much. More complex schemes, involving multistep lookahead, may also be used and generally tend to perform better as the length of the lookahead increases.

Theoretically, our methodology relies on generic cost improvement properties and the superlinear convergence framework of Newton's method, which fundamentally underlies approximation in value space, and related MPC/RL and rollout/policy iteration schemes. A critical requirement of this framework is that the first lookahead step should be executed exactly. This fact has guided our architectural choices, and is apparently an important factor in improving the performance of even the best available chess engines.

\end{abstract}


\vspace{1pc}
\section{Introduction}

The fundamental paper on which computer chess programs are based was written in 1950 by one of the most illustrious modern-day applied mathematicians, C.\ Shannon [Sha50]. It was argued by Shannon that whether the starting chess position is a win, loss, or draw is a question that can be answered in principle through exact minimax search that extends up to game termination with a win by one of the players or a draw.\footnote{The minimax search is simply dynamic programming for minimax control problems, which is used more broadly in optimization and control, in the presence of set uncertainty (see e.g., the books [Ber17], [Ber20], [Ber22a], [Ber23]). In this paper, we do not discuss control problems and the related theoretic issues. However, the ideas of the present paper extend to that context; see [Ber23], Section 2.12.}  However, the answer will probably never be known because this would require extraordinarily long computation. As an alternative, Shannon proposed a limited lookahead of a few moves and evaluating the end positions by means of some scoring function. This is the principle on which all current major computer chess programs are based: they involve a search through a tree, which  is rooted at the current position and extends to a certain  depth. The positions at the leaves of the tree are evaluated using the scoring function, and the move chosen at the root of the tree is the one with best backed-up score. 

The tree may be pruned selectively to save computation time and to extend the length of the search within the given time constraint: this was called  a type B strategy by Shannon, to differentiate it from what he called type A strategy that does not resort to any kind of pruning (except to enhance computational efficiency in executing an exact minimax strategy, as in alpha-beta pruning [KnM75]). Chess-playing computer programs typically use a combination of Shannon's type A and type B strategies, but over the years the balance has shifted towards the type B strategy, with the aim to extend the search to deeper levels. The paper by van den Herik [Her18] provides an overview of the history of computer chess, and notes the contributions of many researchers since Shannon's paper.

Tree search and tree pruning heuristics have been researched extensively in computer chess, and more generally in machine learning, since the 1960s. However, the position evaluation methodology was transformed dramatically in 2017 with the introduction of the AlphaZero chess program (Silver et al.\  [SHS17]), which used a deep neural network to provide a position evaluator that was trained off-line through self-play and self-improvement. This training methodology is couched on the algorithm  of policy iteration, which is fundamental in dynamic programming (DP) and reinforcement learning (RL).\footnote{This is also true of similarly structured programs by Lai [Lai15] for chess, and by Tesauro [Tes94], [Tes95] for backgammon.} It is also couched on Monte Carlo tree search ideas to enhance pruning of the lookahead tree, in the spirit of the type B strategy. The success of AlphaZero was replicated by other chess engines such as Leela Chess Zero; see e.g., the relevant Wikipedia article, and the book by Klein [Kle22]. The Stockfish program, which was defeated by AlphaZero in 2017, was improved substantially later by  modifying its manually-tuned position evaluator to incorporate neural network-based evaluations. The present form of Stockfish is widely viewed as exceptionally strong, and capable of almost perfect chess play. Another very strong engine, which uses similar principles to Stockfish, is Komodo Dragon, the winner of several computer chess world championship tournaments. There is extensive literature and documentation on Stockfish and Komodo Dragon, including open source codes; see e.g., the relevant Wikipedia articles.

In this paper, we propose a new type of computer chess architecture, which  is based on model predictive control (MPC), a methodology that originated in the control system design and optimal control contexts in the 1960s.  There is a strong relation between MPC, and ideas of approximation in value space and rollout (a single policy iteration) that form a major part of RL (see the recent tutorial survey by Bertsekas [Ber24]). Our framework harnesses existing chess engines (with no modification) into a search scheme that involves a combination of Shannon's type A and type B strategies: the MPC portion uses a type A strategy, and the chess engine portion uses whatever type B strategy is built into the engine. We may view our architecture as a meta algorithm that uses traditional chess engines as lower level procedures; hence we adopt the acronym MPC-MC (Model Predictive Control-Meta Chess).\footnote{A {\it meta algorithm} is a broad term that applies to an algorithm that operates on other algorithms, often by modifying or combining them; hence the name ``Meta Chess."}

From a theoretical point of view, MPC-MC allows a synergism of off-line training of the position evaluation function and the on-line search process that is couched on the algorithmic framework of Newton's method. This theoretical view is discussed at length in the recent books by Bertsekas [Ber20], [Ber22a], [Ber23], and the related survey papers [Ber22b] and [Ber24], and will not be explained at any depth in this paper. However, it is fundamentally responsible for the results that we are reporting here. These results suggest that by incorporating any chess engine into our MPC framework, we can improve the performance of that engine, and often dramatically so. As an example, MPC-MC based on Stockfish engines defeats the Stockfish engines on which it is based by overwhelming margins for fast time limits, and by lesser margins for longer time limits (at which the Stockfish play is nearly optimal).

Finally, we note that the structure and characteristics of our architecture apply not only to chess, but also to any two-person zero-sum game, not involving stochastic uncertainty, such as Shogi, Xiangqi, Checkers, Go, Reversi,  etc. It is likely that similar results to the ones reported here can be obtained in the context of these games.

The paper is organized as follows. In the next section we describe our MPC-MC architecture in its one-step lookahead form. In Section 3, we introduce the two main variants of  MPC-MC, which apply to the cases of a deterministic/known  and a stochastic/unknown opponent. We also discuss a so-called fortified variant, which is motivated by experience with rollout algorithms, and is effective against very strong, world champion-caliber opponent engines. In Section 4, we provide detailed experimentation results, which support the theoretical development of Section 3. Moreover, we discuss experiments with another variant of MPC-MC, which uses a ``half-step" lookahead and does not involve a nominal opponent (only a position evaluation for each legal move at the current position, see Section 4.2). Consistent with theoretical predictions, this variant of MPC-MC improves the performance of the position evaluation engine, but not by as much as the one-step lookahead variants. In Section 5, we discuss multistep lookahead versions of MPC-MC, which achieve somewhat better results that their one-step lookahead variants, at the expense of more intensive computation.

\section{The MPC-MC Architecture}

Given a chess position $x$, a typical  chess engine will select a move on the basis of some calculations that involve multistep minimax search (likely approximate, because of pruning of the lookahead tree).\footnote{For a recent example of implementation of a grandmaster-level chess program without search, see  Ruoss et al.\ [RDM24]. This implementation uses a huge-size (270 M parameters) trained neural network position evaluator, and is well suited for use within the MPC-MC architecture; see Section 4.} In particular, it will produce a numerical evaluation $Q(x,u)$ of the position resulting from $x$ after each of the legal moves $u$, and it will play a move $\tl u$ that produces the best evaluation.\footnote{In RL terms, we can view $Q(x,u)$ as the Q-factor of the pair $(x,u)$. We adopt the convention that smaller Q-factors correspond to better moves.} 
Thus, for the purposes of this paper,  a chess engine can be viewed as a function $\m$, which  when faced with a position $x$, plays the move 
$\m(x)=\tl u$, where 
$$\tl u\in\arg\min_{u\in L(x)}Q(x,u),$$
with $L(x)$ denoting the set of legal moves at $x$.
The function 
$$E(x)=\min_{u\in L(x)}Q(x,u).$$
will be called the {\it evaluation function} of the engine; it provides a numerical evaluation of any given position $x$ (castling and record keeping to detect a drawn chess position, such as a three-move repetition, are incorporated as part of the position).\footnote{For some engines the formula for $E(x)$ is  not strictly correct, because some of the legal moves at $x$ are pruned, so the minimization defining $E(x)$ is approximate.} 
 In this paper, we assume that an engine is memoryless, so that $Q(x,u)$ depends only on the pair $(x,u)$ and not on the earlier game calculations.\footnote{Some chess engines are not truly memoryless; for example they build hash tables of evaluated positions, which are carried from one move to the next in the course of a game. The effect of engine memory on the performance of MPC-MC is a complex issue that has yet to be fully assessed.} 

To play against an opponent, our MPC-MC architecture also selects a move in response to a given position. However, to compute this move it {\it uses two engines\/}, such as Stockfish (SK for short), Komodo Dragon (KD for short),  Leela Chess Zero (LC0 for short), etc. 

\begin{itemize}

\item[(a)] The {\it position evaluator\/}. This is a chess engine, which produces a numerical evaluation of any given position.

\item[(b)] The {\it nominal opponent\/}. This is either an exact replica or an approximation to the true opponent (engine or human), whom we intend to play against. It outputs deterministically a move to play at any given chess position. In the absence of knowledge of the true opponent, a reasonable choice is to use a competent chess engine as nominal opponent, such as for example the one used to provide position evaluations.\footnote{We will argue later that it is important {\it not to use a relatively poor nominal opponent\/}, which would lead us to underestimate the true opponent.}
\end{itemize}

The nominal opponent and the position evaluator engines may be different. Moreover, the nominal opponent engine may be changed from game to game to adapt to the true opponent at hand. Note that stored knowledge of the evaluator and nominal opponent  engines, such as an opening book or an endgame database, are indirectly incorporated into the MPC-MC player. 

To describe mathematically the move selection of MPC-MC at a position $x_k$, and to make the connection with the optimal control and MPC/RL framework, we use the following notation:
\begin{itemize}
\item[$\bullet$] $x_k$ is the chess position at time/move $k$.
\item[$\bullet$] $u_k$ is a legal move at time $k$ in response to position $x_k$.
\item[$\bullet$] $w_k$ is the move choice of the nominal opponent at time $k$ in response to position $x_k$ followed by move $u_k$.
\end{itemize}
The resulting position at time $k+1$ is given by
$$x_{k+1}=f(x_k,u_k,w_k),$$
where $f$ is a known function. This corresponds to a dynamic system in the standard MPC framework, where $x_k$ is viewed as the state, $u_k$ is viewed as the control, and $w_k$ is viewed as a known or unknown (possibly random) disturbance.

\begin{figure}[ht]
\captionsetup{singlelinecheck=off}
\begin{center}
\centerline{{\includegraphics[width=1.0\columnwidth]{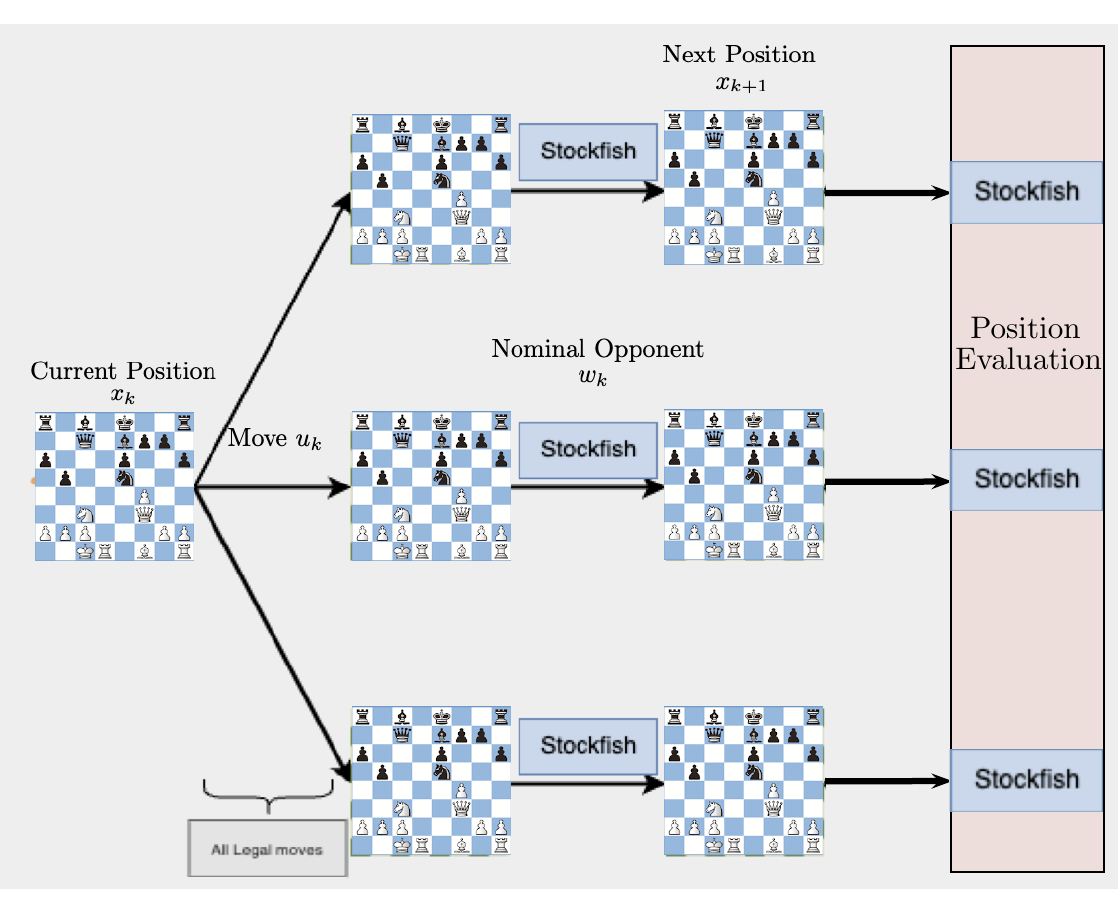}}}
\caption{\small Schematic illustration of the MPC-MC framework with one-step lookahead. The position evaluator and the nominal opponent in the figure are SK engines.}
		\label{figonestepdet}
		\end{center}
\end{figure}

The structure of the MPC-MC architecture with one-step lookahead is shown in Fig.\ \ref{figonestepdet}. Here is the sequence of calculations by which it selects a move $u_k$ at a given position $x_k$:

\begin{itemize}
\item[(1)] We generate all legal moves $u_k$ at $x_k$.

\item[(2)] For each pair $(x_k,u_k)$, we use the nominal opponent engine to  evaluate the position $(x_k,u_k)$ and to generate the corresponding best move 
$\nu(x_k,u_k)$,
where $\nu$ is a given function.

\item[(3)] For each $u_k$, we use the position evaluator engine to evaluate the position 
$$x_{k+1}=f\big(x_k,u_k,\tl w_k\big)$$ 
corresponding to the nominal opponent move 
$$\tl w_k=\nu(x_k,u_k).$$

\item[(4)] We select the move $u_k$ that corresponds to the position $x_{k+1}$ that has best evaluation.
\end{itemize}

Note that to select a move, MPC-MC requires a total of roughly $m$ nominal opponent move generations and $m$ position evaluations, where $m$ is a typical number of legal moves at $x$. Thus, MPC-MC requires roughly $2m$ times more computation than a single position evaluation by one of the engines. 
However, it is important to note that MPC-MC is very well suited for parallelization, and given sufficient parallel computation resources, it roughly requires only twice the amount of computation time of a single engine evaluation.\footnote{If the nominal opponent and position evaluator engines use parallelization for search, then MPC-MC uses implicitly the same parallelization. However, MPC-MC uses additional parallelization: at each stage of lookahead, multiple nominal opponent moves and positions evaluations can be executed in parallel, with nearly 100 percent efficiency.}

\section{Deterministic and Stochastic Variants of MPC-MC}

We will now discuss two basic variants of our MPC-MC architecture. In the first variant, called {\it deterministic\/}, we can predict exactly the response of the true opponent, and use this prediction as the nominal opponent move. This variant can be viewed as a special case of a standard MPC architecture.  In the second variant, called {\it stochastic\/}, we can generate only an approximate prediction. This variant can be viewed as an approximation to an MPC architecture.

\subsection*{Deterministic MPC-MC}
In the deterministic variant of MPC-MC, we are able to predict exactly the move of the true opponent in response to a position $x_k$ and legal move $u_k$, and use the prediction as the nominal opponent move. We thus assume that the nominal opponent engine replicates the play of the true opponent. In this case, if $\nu(x,u)$ denotes the nominal opponent move in response to $x$ followed by $u$, the sequence of positions generated during the actual game evolves according to
\begin{equation} \label{eq-dynsys1}
x_{k+1}=F(x_k,u_k),
\end{equation}
where 
\begin{equation} \label{eq-dynsys2}
F(x_k,u_k)=f\big(x_k,u_k,\nu(x_k,u_k)\big).
\end{equation}
Here $\nu(x_k,u_k)$ is the move generated by both nominal and true opponent engines in response to position $x_k$ and MPC-MC moves $u_k$.

In optimal control terms, Eqs.\ (\ref{eq-dynsys1})-(\ref{eq-dynsys2}) represent a chess game as the evolution of a deterministic controlled system with state $x$ and control $u$. Regarding the cost function, a nonzero cost is incurred only at the terminal positions where one of the opponents wins the game. The optimal cost function $J^*(x)$ is the solution of the Bellman equation, the fundamental equation of exact DP:
\begin{equation} \label{eq-bellman}
J^*(x)=\min_{u\in L(x)}J^*\big(F(x,u)\big),
\end{equation}
with
$$J^*(x)\ne0$$
for all positions $x$ where one of the players can win, regardless of the play of the other player, and with  
$$J^*(x)=0$$
for all other positions $x$, which are theoretical draws. 

Of course the optimal cost function $J^*$ is unknown at the starting chess position (and most other positions),\footnote{Many grandmasters believe that the starting position $s$ is a theoretical draw [$J^*(s)=0$], based on the near perfect play of top engines, which are virtually unbeatable by other engines (let alone humans) when the time limit per move is relatively long.} and is unlikely to be calculated in the near future, in view of the complexity of the chess game. In the approximation in value space approach of RL, which is very similar to the MPC methodology, we approximate the unknown function $J^*$ with an approximation. Accordingly, in the MPC-MC architecture, $J^*$ is approximated by the evaluation function $E$ of the position evaluator engine. In particular, at position $x$, the MPC-MC player selects the move
\begin{equation} \label{eq-opt}
\tl u\in\arg\min_{u\in L(x)}E\big(F(x,u)\big),
\end{equation}
where the needed values of $F$ are calculated by using the nominal opponent engine (also true opponent), and the needed values of $E$ are calculated by the position evaluator engine.  

An important fact is that the performance of the MPC-MC player is better than the performance of the position evaluator engine, provided the evaluator engine's play is relatively close to optimal. This follows from a theoretical framework, which applies more broadly than chess, and  explains the performance  of MPC and RL schemes that are based on approximation in value space. The development, justification, and visualization of this framework is the focal point of the books [Ber20], [Ber22a], [Ber23], and the survey papers [Ber22b] and [Ber24], which were noted earlier. The performance improvement property also relates to Newton's method applied to the Bellman Eq.\ (\ref{eq-bellman}). 

In this paper, we will not discuss further the theoretical backdrop just sketched, but instead we will demonstrate experimentally in Section 4 how the performance of the MPC-MC player is superior to the performance of its position evaluator engine. This performance improvement is generally significant, but tends to diminish in absolute terms as the position evaluator engine approaches optimality; after all if the position evaluator engine plays perfect/optimal chess, its performance cannot be improved. However, in theory, the performance improvement of the MPC-MC player increases in {\it relative terms}  as the position evaluator engine approaches optimality. In particular, we have
\begin{equation} \label{eq-suplinear}
{J_{\tl \m}(x)-J^*(x)\over \tl J(x)-J^*(x)}\to0,\qquad \hbox{as \ } \tl J\to J^*,
\end{equation} 
where $\tl \m$ represents the move selection policy of the MPC-MC player, $J_{\tl \m}(x)$ is its performance starting from any position $x$, and $\tl J(x)$ is the position evaluation, as given by the position evaluator engine.\footnote{For the superlinear relation (\ref{eq-suplinear}) to make sense, $J_{\tl \m}(x)$ and $\tl J(x)$ must be suitably normalized to be comparable to $J^*$. A popular normalization convention is to set $J^*(x)=0$ for a loss, $J^*(x)=1$ for a win, and to view $\tl J(x)$ as a ``probability" of winning starting from the current position $x$.} This is a superlinear performance improvement relation that is typical of Newton's method; see the sources cited earlier. Our experimental results are consistent with this relation.

We finally note that the use of the position evaluator engine to approximate the optimal cost function is reminiscent of the rollout algorithm from DP/RL, whereby the optimal cost function is approximated by the cost function of some policy known as the {\it base policy\/}; see the sources cited earlier for an extensive account of rollout algorithms, and for many references to their remarkable effectiveness. In our case, the base policy is the move selection policy of the position evaluator engine, and its cost function is approximated by the corresponding engine evaluations.

\subsection*{Stochastic MPC-MC}

In the stochastic variant of MPC-MC, we cannot predict exactly the move of the true opponent in response to $(x_k,u_k)$, and we use instead the move generated by a nominal opponent engine that approximates the play of the true opponent. Again, if $\nu(x,u)$ is the nominal opponent move in response to $x$ followed by $u$, the MPC-MC player calculates its move according to
\begin{equation} \label{eq-opt1}
\tl u\in\arg\min_{u\in L(x)}E\big(F(x,u)\big),
\end{equation}
cf.\ Eq.\ (\ref{eq-opt}),
where 
\begin{equation} \label{eq-dynsys2a}
F(x,u)=f\big(x,u,\nu(x,u)\big).
\end{equation}
However, the positions generated during the actual game will occasionally deviate from the positions generated by the equation
\begin{equation} \label{eq-dynsys1a}
x_{k+1}=F(x_k,u_k),
\end{equation}
because the play of the true opponent can deviate from the play of the nominal opponent.

While the performance improvement property of Eq.\ (\ref{eq-suplinear}) cannot be established for the stochastic MPC-MC player, it evidently holds approximately, provided that  the nominal opponent is a strong player. In fact our experiments indicate that the nominal opponent engine should play at least as well or better than the true opponent. As an explanation of why we need a strong nominal opponent, we note that  the MPC-MC architecture may select a poor and even catastrophic move $u_k$ because  the nominal opponent produces a poor response $\nu(x_k,u_k)$, leading to $x_{k+1}$ which is favorably judged by the position evaluator engine. It is thus essential that the nominal opponent does not underestimate the true opponent.

\subsection*{Fortified Move Generation}

While the MPC-MC architecture has worked well in our computational experiments, it may make occasional errors, even when it uses very strong nominal opponent and position evaluator engines. In particular, we have observed rare mistakes, which seem to be due to the approximation of minimax play of the opponent with the moves produced by the nominal opponent. This type of situation is common in {\it truncated} rollout algorithms for general approximate DP/RL settings, where the search with a base policy is extended to only a limited depth of lookahead. A useful supplement to truncated rollout algorithms is {\it fortification\/}, whereby we follow the base policy at states where the rollout policy appears to be ineffective (we refer to sources given earlier, such as the books [Ber19] and [Ber20], for a detailed discussion).

We have thus considered an MPC-MC fortification strategy, which works as follows.  For a given position $x_k$, once a move $\tl u_k$ is computed by using the MPC-MC policy of Eq.\ (\ref{eq-opt}) or Eq.\ (\ref{eq-opt1}), it is compared with the move suggested by the position evaluator  at $x_k$, call it $\hat u_k$. The MPC-MC fortification strategy then is to play $\hat u_k$ if its evaluation $Q(x_k,\hat{u}_k)$ is better than the one of $\tl u_k$, and to play $\tl u_k$ otherwise. 

As the preceding discussion suggests, the fortification strategy is conservative, but provides some safeguards against overambitious play by the (unfortified) MPC-MC architecture. Our computational experiments, given in the next section, indicate that fortification is effective against very strong opponents, such as powerful SK engines, which can exploit even small errors by MPC-MC. On the other hand, fortification may lead to a relatively small performance degradation against weak opponents, against which MPC-MC has an overwhelming advantage.

\section{Computational Results}
In this section, we will present our computational results for the MPC-MC architecture applied to two different types of chess engines. We will first use SK and/or KD chess engines of varying strengths as the nominal opponent and the position evaluator.\footnote{The preliminary Python implementation of MPC-MC with one-step lookahead, and the SK and KD engines, can be downloaded at \url{https://yuchaotaigu.github.io/research/MPC_MC_1step.ipynb}.} These engines are specially designed for playing chess, and they are incorporated into the MPC-MC architecture without modification. The strongest versions of these engines have won prominent computer chess competitions in recent years. Still, our results show that MPC-MC with one-step lookahead can improve their performance. Moreover, our results indicate that fortification in MPC-MC is beneficial against very strong opponents.

We have also used the chess engines developed by Ruoss et al [RDM24] (at Google DeepMind) within the MPC-MC architecture. These are engines that rely exclusively on (off-line trained) transformer neural networks to provide position evaluations {\it without further search\/}. At the current position and for each legal move, they calculate a Q-factor and then they choose the move with best Q-factor. These engines attain a strong grandmaster-level performance against human opponents, but are  generally weaker than the SK and KD engines. Still, in a limited set of experiments, we have verified that MPC-MC can provide a significant performance improvement, consistently with our experience with the SK and KD engines.\footnote{We have used these engines for only part of a game. In particular, the first 12 moves of each game were generated using SK (playing against itself) in order to reach a variety of middlegame positions from which to start using the MPC-MC architecture.  Also, because the transformer engines have not been trained to handle endgames well, we have substituted instead the SK engine once the number of pieces reached 12 or less.}  Note that the transformers that we used in these experiments have similar structure with versatile large language models, which can be used in a broad range of applications. This suggests that we can expect offline-trained transformers to be well-suited for use within our MPC-based architecture in a variety of settings, which involve minimax sequential decision making (including two-person zero-sum games other than chess). Further research in this direction appears to be promising. 

\subsection{MPC-MC with Stockfish and Komodo Dragon Chess Engines}

We will first provide some computational details of the MPC-MC architecture with one-step lookahead, when the specialized chess engine Stockfish (SK) and/or the freely available version of Komodo Dragon (KD) are used as nominal opponent and/or position evaluator. We will not repeat the generic procedures of MPC-MC introduced in Section~3. Instead, we will focus on describing how the engines can be incorporated into MPC-MC, to play against themselves or against each other. We will then describe the setup of our computational studies, where we test both the deterministic and the stochastic MPC-MC architectures, and evaluate the effect of fortification. 

Given the chess position $x_k$ at time $k$ and a legal move $u_k$, both the SK and KD engines select a move $\nu(x_k,u_k)$ via some built-in function. In addition, these engines can be used to provide scalar evaluations $E(x_{k+1})$ of any chess position $x_{k+1}$ with some additional calculation. Both the quality of move selection and the position evaluation can be affected by a variety of factors. In our tests, we have relied solely on the computational time limit to affect the performance of these engines. In particular, given two engines of the same type (be it SK or KD), we view the one with higher time limit as ``stronger."

We have tested both the deterministic and stochastic MPC-MC architecture, where the engines' time limits are set to $0.5$, $2$, and $5$ seconds (at 5 secs, the play of KD is very strong, while the play of SK is nearly perfect). To eliminate the effect of stored hash tables on the performance, only engines without  stored data are used in our computation. In the deterministic version of MPC-MC, the moves selected by the nominal opponent corresponding to the selected rollout moves are played in the actual game. In the stochastic MPC-MC version, the moves of the true opponent are selected by an independent engine that has the same strength as the nominal opponent. Due to various reasons, engines with the same strength and no hash tables can still select different moves given the same pair $(x_k,u_k)$. Therefore, the nominal opponent may predict moves that are different from those selected by the true opponent, which makes the move selected by MPC-MC a stochastic one. Since the engines also take advantage of available computational resources effectively, for fairness we have limited the computing resources of the true opponent to be equal to that used for evaluating one legal move in MPC-MC. 

Our test results for deterministic MPC-MC are summarized in Table~\ref{tabel:det}, where we considered three different engine configurations. In the games listed in the column ``SK vs SK," both the nominal opponent, the position evaluator, and the true opponent are SK. Similarly, the nominal opponent, the position evaluator, and the true opponent are KD for those listed in the ``KD vs KD" column. For the games in the column ``SK vs KD," the nominal and the true opponent are KD, while the position evaluator is SK.
 
 In each setting, we tested MPC-MC (results of which are listed under ``Std.") and its fortified variant (under ``Fort."), where a win, a draw, or a loss count for $1$, $0.5$, and $0$ points, respectively. In all the games we tested, MPC-MC did not lose a single game, and the performance improvement is most clear when the engines used are relatively weak. When competing against a very strong opponent, such as SK at time limit 5 secs, our results show that fortification can be beneficial. Note that the freely available version of KD that we used is a weaker engine than SK at the same time limit, and our computational results bear this out.

\begin{table}[h!]
\centering
\caption{Test Results for Deterministic MPC-MC}
\label{tabel:det}
\begin{tabular}{|c|c|c|c|c|c|c|c|c|c|}
\hline
\multirow{2}{*}{\textbf{Strength (secs)}} & \multicolumn{2}{c|}{\textbf{SK vs SK}} & \multicolumn{2}{c|}{\textbf{KD vs KD}} & \multicolumn{2}{c|}{\textbf{SK vs KD}} \\ \cline{2-7} 
 & \textbf{Std.} & \textbf{Fort.} & \textbf{Std.} & \textbf{Fort.} & \textbf{Std.} & \textbf{Fort.} \\ \hline
0.5 & 7.5-2.5 & 8-2 & 8.5-1.5 & 8-2 & 10-0 & 10-0 \\ \hline
2 & 5-5 & 5.5-4.5 & 6.5-3.5 & 8.5-1.5 & 9.5-0.5 & 9-1 \\ \hline
5 & 5-5 & 5.5-4.5 & 6.5-3.5 & 7.5-2.5 & 10-0 & 9-1 \\ \hline
\end{tabular}
\end{table}

Unlike a deterministic MPC-MC, its stochastic counterpart cannot predict the moves of true opponent exactly. In our stochastic  MPC-MC tests, the nominal and the true opponent were represented by two engines of the same strength, which, however, for reasons of internal implementation, can select different moves given the same position.
Our computational results are shown in Table~\ref{table:sto}, where again MPC-MC did not lose a single game, as in the deterministic case. Moreover, the scores obtained also lead to similar conclusions as those in deterministic MPC-MC. Since MPC-MC is  strongly favored in the games ``KD vs KD" and ``SK vs KD," we did not test the fortified variant for these cases.  

\begin{table}[h!]
\centering
\caption{Test Results for Stochastic MPC-MC}
\label{table:sto}
\begin{tabular}{|c|c|c|c|c|c|c|c|c|c|}
\hline
\multirow{2}{*}{\textbf{Strength (secs)}} & \multicolumn{2}{c|}{\textbf{SK vs SK}} & \multicolumn{2}{c|}{\textbf{KD vs KD}} & \multicolumn{2}{c|}{\textbf{SK vs KD}} \\ \cline{2-7} 
 & \textbf{Std.} & \textbf{Fort.} & \textbf{Std.} & \textbf{Fort.} & \textbf{Std.} & \textbf{Fort.} \\ \hline
0.5 & 8-2 & 7-3 & 7-3 & NA & 8-2 & NA \\ \hline
2 & 5.5-4.5 & 6.5-3.5 & 6.5-3.5 & NA & 8-2 & NA \\ \hline
5 & 10-10 & 10.5-9.5 & 6-4 & NA & 9-1 & NA \\ \hline
\end{tabular}
\end{table}

\subsection{Comparison with a Half-Step Lookahead Version of MPC-MC}

\begin{figure}[ht!]
\captionsetup{singlelinecheck=off}
\begin{center}
\centerline{{\includegraphics[width=1.0\columnwidth]{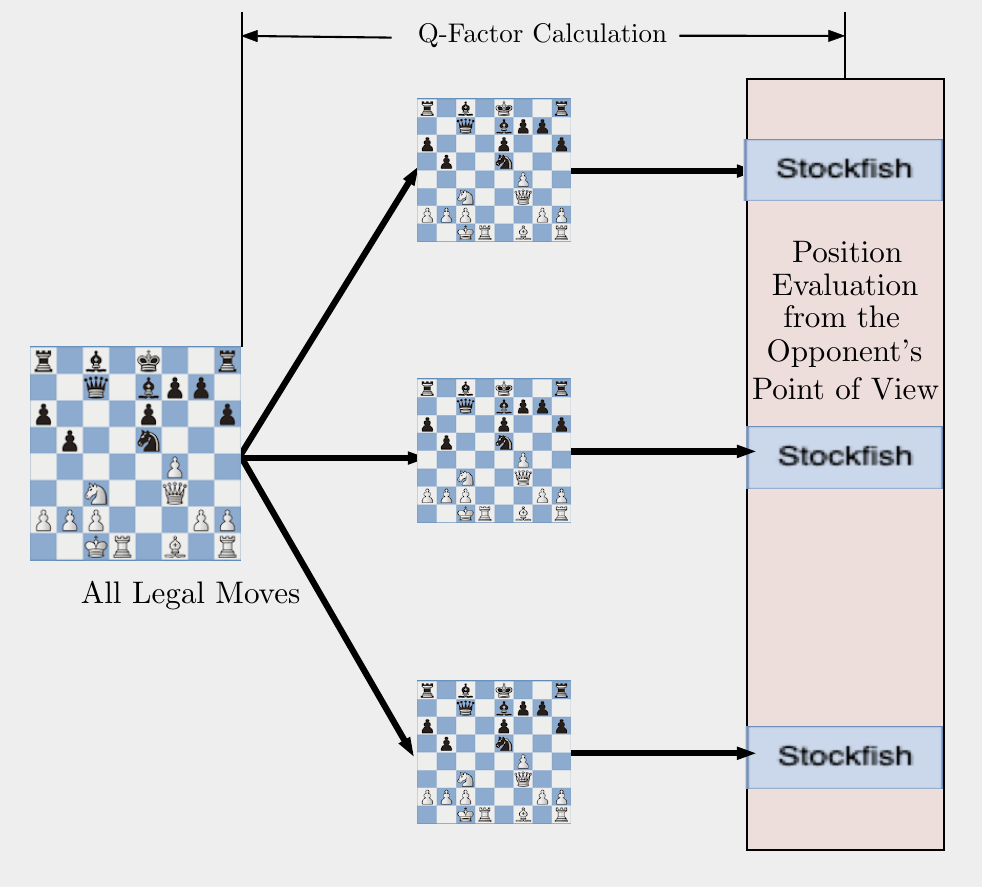}}}
\caption{\small Schematic illustration of the MPC-MC framework with half-step lookahead. Here the SK engine is used to evaluate (from the opponent's point of view) all legal moves at the current position. Then MPC-MC selects the move that is most difficult for the opponent, i.e., the one that results in the worst evaluation according to SK.}
		\label{fighalfstep}
		\end{center}
\end{figure}

Let us also consider a simpler version of MPC-MC, which does not use a nominal opponent engine, only a position evaluator. We call this the {\it half-step lookahead version\/}, and we illustrate it in Fig.\ \ref{fighalfstep}.  Here the evaluator engine  considers all legal moves at the current position, and evaluates them from the point of view of the opponent. The half-step version of MPC-MC then selects  the legal move that results in the position that is worst from the opponent's point of view. 

Theoretically, this version also has a Newton step interpretation, but the Newton step is somewhat less reliable. The reason has to do with the lack of concavity of the minimax form of the Bellman operator (see [Ber22a], Section 3.9, [Ber22c], Chapter 5).
We will not go into further details, but we will instead discuss briefly our computational results. These results  are favorable but not as favorable as the ones we presented earlier for MPC-MC with one-step lookahead against the SK engine.
In particular, 
 half-step MPC-MC (using the SK evaluator at  0.5 sec) vs SK won a ten-game match by 6.5-3.5, while for one-step MPC-MC, the corresponding result was 8-2.
Against SK at 2 sec, half-step MPC-MC drew a ten-game match (5-5), while for one-step MPC-MC, the result was 5.5-4.5; see the left side of Table \ref{table:half}. 

Similar results have been obtained when using the chess engines developed by Ruoss et al [RDM24] as position evaluators.\footnote{The paper by Ruoss et al [RDM24] provides three different chess engines based on transformer neural networks, representing a Q-factor evaluation, a position evaluation, and a policy, respectively. We have used the one that provides a Q-factor evaluation. } In particular, in 10-game matches with 136 M and 270 M transformers, MPC-MC won by scores of 7.0-3.0 and 7.5-2.5, respectively (with one loss in each case); see the right side of Table \ref{table:half}.

\begin{table}[ht!]
\centering
\caption{Test Results for MPC-MC with Half-Step Lookahead. The  two left columns represent 10-game match  results for playing MPC-MC with SK as position evaluator against SK.  The  two right columns represent results for playing MPC-MC with the transformer-based engine by Ruoss et al [RDM24] (abbreviated TF) as position evaluator against TF.}
\label{table:half}
\begin{tabular}{|c|c||c|c|}
\hline
\textbf{Strength} & \textbf{SK vs SK} & \textbf{Strength} & \textbf{TF vs TF} \\ \hline
0.5 secs& 6.5-3.5 & 136 M & 7-3  \\ \hline
2 secs& 5-5 & 270 M& 7.5-2.5 \\ \hline
\end{tabular}
\end{table}

Note that the half-step lookahead version of MPC-MC can also be viewed as a variant of the evaluator engine, which involves deeper lookahead by a half move. On the other hand, the lookahead of strong engines is very long (over 20), so the computational results that we have presented cannot be explained merely by the lengthening of the lookahead by a half move. Our speculation is that making the initial  half-step lookahead exact (without any pruning) implements a Newton iteration for solving the  Bellman equation  associated with the underlying minimax problem, and is likely responsible for our favorable computational results.

\section{MPC-MC with Multistep Lookahead}

\begin{figure}[ht!]
\captionsetup{singlelinecheck=off}
\begin{center}
\centerline{{\includegraphics[width=1.0\columnwidth]{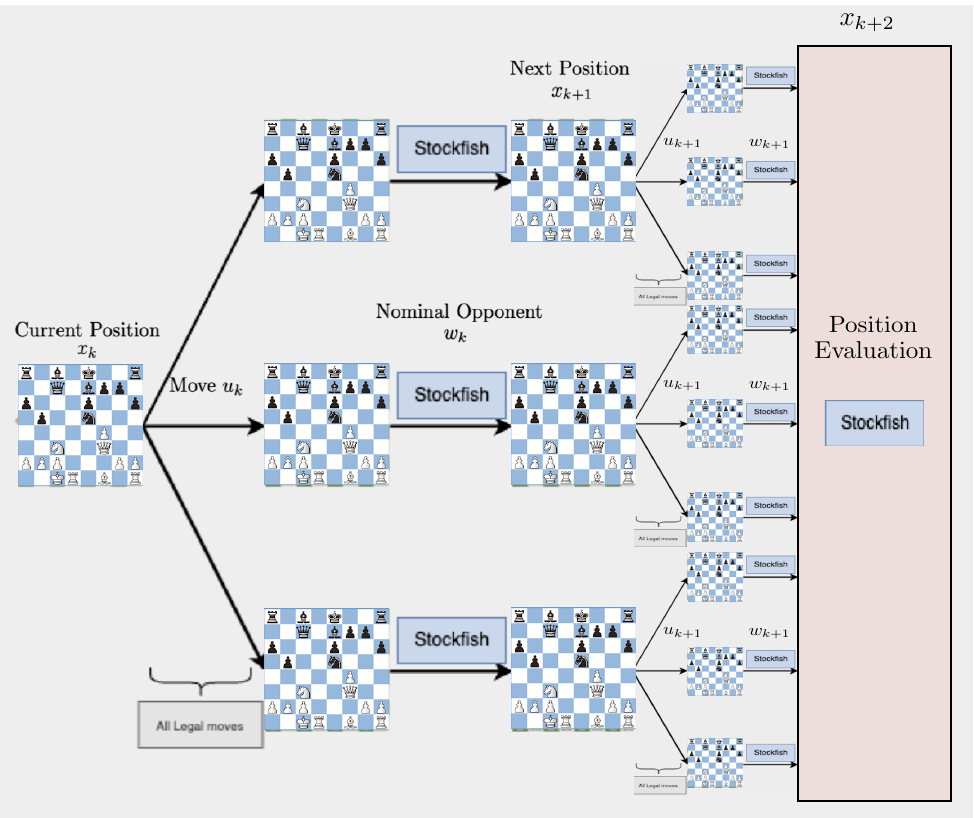}}}
\caption{\small Schematic illustration of the MPC-MC framework with two-step lookahead, and a deterministic nominal opponent. The position evaluator and the nominal opponent in the figure are SK engines.}
		\label{figtwostepdet}
		\end{center}
\end{figure}

We next consider the structure of the MPC-MC scheme with two-step lookahead (the case of lookahead longer than two is similar); see Fig.\ \ref{figtwostepdet}:

\begin{itemize}
\item[(1)] At  the current position $x_k$, we generate all legal moves $u_k$ (say $m$ in number).

\item[(2)] For each pair $(x_k,u_k)$, we use the nominal opponent to generate a single best move $w_k$, resulting in the position 
$$x_{k+1}=f(x_k,u_k,w_k).$$

\item[(3)] For each of the resulting positions $x_{k+1}$, we generate all legal moves $u_{k+1}$.

\item[(4)] For each pair $(x_{k+1},u_{k+1})$, we use the nominal opponent to generate a single best move $w_{k+1}$, resulting in the position 
$$x_{k+2}=f(x_{k+1},u_{k+1},w_{k+1}).$$

\item[(5)] We evaluate each of the possible positions $x_{k+2}$ by using the position evaluator engine, and select as next move $u_k$ the one that leads to the position $x_{k+2}$ with best evaluation.
\end{itemize}

Thus, there are roughly a total of at most $m^2$ position evaluations and $m^2+m$ nominal opponent move generations, where $m$ is representative of the number of possible legal moves at the positions that may result from the two-step lookahead process. These calculations can be expedited by using parallel computation, but they can also be expedited by pruning the tree that corresponds to the second level of lookahead. One way to do this is to consider the  evaluations of the positions $x_{k+1}$ following the first layer of nominal opponent moves $w_k$. Then prune some of the less promising positions  $x_{k+1}$, i.e., those $x_{k+1}$ that have relatively low evaluation, as provided by the local opponent engine. This is consistent with MPC theory, which suggests that it is important to execute exactly the first level of lookahead minimization, but not nearly as accurately for the second and subsequent levels of lookahead (see the books [Ber20], [Ber22a], and [Ber23] cited earlier). Some testing results of MPC-MC with two-step lookahead for ``SK vs SK" are summarized in Table~\ref{table:2stp}. Comparing with the corresponding results of Tables~\ref{tabel:det} and \ref{table:sto}, it appears that longer lookahead indeed produces a performance improvement, as expected. We are planning a more extensive evaluation that will allow us to reach reliable quantitative conclusions in this regard.

\begin{table}[h!]
\centering
\caption{MPC-MC with Two-Step Lookahead}
\label{table:2stp}
\begin{tabular}{|c|c|c|}
\hline
\textbf{Strength (secs)} & \textbf{Deterministic} & \textbf{Stochastic} \\ \hline
0.5 & 6-0 & 5.5-0.5\\ \hline
2 & 1.5-0.5 & 1-1  \\ \hline
\end{tabular}
\end{table}

\section{Concluding Remarks}

The ideas of this paper were motivated by theoretical insights from the MPC, RL, approximation in value space, and rollout methodologies. These ideas center around Newton's method for solving Bellman's equation associated with an underlying DP problem, and apply very generally to discrete and continuous spaces sequential decision problems (see the books and papers noted earlier). They apply to computer chess in particular, after its minimax two-player structure is changed to a one-player sequential decision structure through the use of the nominal opponent engine. Consistent with the theoretical insights, we have verified that the MPC-MC architecture provides a boost in the performance of existing chess engines, including engines that play near perfect chess, like Stockfish at a 5 secs per move time limit.

We distinguished between a deterministic architecture, where the nominal opponent replicates exactly the actual opponent, and a stochastic architecture, where it does not. An interesting observation from our experiments is that the performance difference between the deterministic and stochastic versions of MPC-MC is relatively small, as long as the nominal opponent does not significantly underestimate the strength of the actual opponent. We believe that multistep lookahead will provide an additional boost in performance, but this remains to be confirmed with additional testing.

While each move of MPC-MC requires many position evaluations with the nominal opponent and position evaluation engines, these evaluations can be done in parallel with near perfect efficiency (twice as long engine time for one-step lookahead and three times as long for two-step lookahead, given enough parallel computing resources). This was verified approximately with the use of cloud-based parallel computations, and we expect that it can be verified using dedicated parallel computing hardware, as well.

We finally note that the structure of the MPC-MC architecture can be applied to a host of other two-person games for which computer engines are available. Key ideas in this regard are: 

\begin{itemize}
\item The first step of lookahead is done exactly (all legal moves should be considered in the context of computer chess and MPC-MC). 
\item The actual opponent is replaced by a nominal opponent, in order to apply one-player MPC methods.
\end{itemize}

These ideas apply more generally to deterministic minimax problems with arbitrary state space and finite control space; see the textbook [Ber23], Section 2.12. Further research will be necessary to explore the potential of the ideas of this paper within this broader domain.

\def\refspace{\par\noindent}
\section*{References}

\refspace[Ber17] Bertsekas, D.\ P., 2017.\ Dynamic Programming and Optimal Control,  Vol.\ I, 4th Edition, Athena Scientific, Belmont, MA.

\refspace[Ber19] Bertsekas, D.\ P., 2019.\ Reinforcement Learning and Optimal Control,  Athena Scientific, Belmont, MA.

\refspace[Ber20] Bertsekas, D.\ P., 2020.\ Rollout, Policy Iteration, and Distributed Reinforcement Learning,  Athena Scientific, Belmont, MA.

\refspace[Ber22a] Bertsekas, D.\ P., 2022.\ Lessons from AlphaZero for Optimal, Model Predictive, and Adaptive Control,  Athena Scientific, Belmont, MA  (can be  downloaded from the author's website).

\refspace[Ber22b] Bertsekas, D.\ P., 2022.\ ``Newton's Method for Reinforcement Learning and Model Predictive Control," Results in Control and Optimization, Vol.\ 7, pp.\ 100-121.

\refspace[Ber22c] Bertsekas, D.\ P., 2022.\ Abstract Dynamic Programming, 3rd Edition, Athena Scientific, Belmont, MA  (can be  downloaded from the author's website).

\refspace[Ber23] Bertsekas, D.\ P., 2023.\ A Course in Reinforcement Learning,  Athena Scientific, Belmont, MA  (can be  downloaded from the author's website).

\refspace[Ber24] Bertsekas, D.\ P., 2024.\ ``Model Predictive Control, and Reinforcement Learning: A Unified Framework Based on Dynamic Programming," arXiv preprint arXiv:2406.00592; also in Proc.\ IFAC NMPC.

\refspace[Her18] van den Herik, H.\ J., 2018.\ ``Computer Chess: From Idea to DeepMind," ICGA Journal, Vol.\ 40, pp.\ 160-176.
 
\refspace[KnM75] Knuth, D.\ E., and Moore, R.\ W., 1975.\ ``An Analysis of Alpha-Beta Pruning,"
Artificial Intelligence, Vol.\ 6, pp.\ 293-326.

\refspace[Lai15] Lai, M., 2015.\ ``Giraffe: Using Deep Reinforcement Learning to Play Chess," arXiv preprint arXiv:1509.01549.

 \refspace[RDM24] Ruoss, A., Del\'etang, G., Medapati, S., Grau-Moya, J., Wenliang, L.\ K., Catt, E., Reid, J., and Genewein, T., 2024.\ ``Grandmaster-Level Chess Without Search," arXiv:2402.04494.

\refspace[SHS17] Silver, D., Hubert, T., Schrittwieser, J., Antonoglou, I., Lai, M., Guez, A., Lanctot, M., Sifre, L., Kumaran, D., Graepel, T., and Lillicrap, T., 2017.\ ``Mastering Chess and Shogi by Self-Play with a General Reinforcement Learning Algorithm,"  arXiv:1712.01815.

\refspace [Sha50] Shannon, C., 1950.\  ``Programming a Digital Computer for Playing
Chess," Phil.\ Mag., Vol.\ 41, pp.\ 356-375.

\refspace[Tes94] Tesauro, G.\ J., 1994.\ ``TD-Gammon, a Self-Teaching 
Backgammon Program, Achieves Master-Level Play,'' Neural Computation, Vol.\ 6, pp.\ 
215-219.

\refspace[Tes95] Tesauro, G.\ J., 1995.\ ``Temporal Difference Learning and TD-Gammon,'' Communications of the ACM, Vol.\ 38, pp.\ 58-68.

\end{document}